\documentclass[11pt,letterpaper]{article}
\pdfoutput=1
\usepackage{emnlp2016}
\usepackage{times}
\usepackage{latexsym}

\emnlpfinalcopy

\def\emnlppaperid{***}

\usepackage{xspace}
\usepackage{booktabs}
\usepackage{multirow}
\usepackage{graphicx}
\usepackage{url}
\makeatletter
\newcommand{\@BIBLABEL}{\@emptybiblabel}
\newcommand{\@emptybiblabel}[1]{}
\makeatother
\usepackage[hidelinks]{hyperref}

\usepackage{amsmath}
\usepackage{amsfonts} 
\usepackage{amssymb} 

\usepackage[american]{babel}

\usepackage{tablefootnote}

\newcommand{\KL}[2]{KL[#1\|#2]}
\newcommand\geocorpus{{{\sc Geo} corpus}\xspace}
\newcommand\geoquery{{\sc GeoQuery}\xspace}
\newcommand\maze{{\sc SAIL}\xspace}
\newcommand\nlmaps{{\sc NLmaps}\xspace}
\newcommand\stos{{\sc S2S}\xspace}
\newcommand\xzyx{{\sc Seq4}\xspace}
\newcommand\xzyxqlq{{\sc Seq4}\xspace}

\title{Semantic Parsing with Semi-Supervised Sequential Autoencoders}

\author{Tom\'a\v s Ko\v cisk\'y$^{\dag\ddag}$ \quad
  G\'abor Melis$^\dag$ \quad
  Edward Grefenstette$^\dag$ \\
  \textbf{Chris Dyer$^\dag$ \quad
  Wang Ling$^\dag$ \quad
  Phil Blunsom$^{\dag\ddag}$ \quad Karl Moritz Hermann$^\dag$}\\
  $^\dag$Google DeepMind \quad $^\ddag$University of Oxford\\
  {\tt \{tkocisky,melisgl,etg,cdyer,lingwang,pblunsom,kmh\}@google.com}
  }

\date{}

\begin{document}

\maketitle

\begin{abstract}
  We present a novel semi-supervised approach for sequence transduction and
  apply it to semantic parsing. The unsupervised component is based on a
  generative model in which latent sentences generate the unpaired logical
  forms.
  We apply this
  method to a number of semantic parsing tasks focusing on domains with limited
  access to labelled training data and extend those datasets with synthetically
  generated logical forms.
\end{abstract}

\section{Introduction}

Neural approaches, in particular attention-based sequence-to-sequence models,
have shown great promise and obtained state-of-the-art performance for sequence
transduction tasks including machine translation \cite{bahdanau:2014:nmt},
syntactic constituency parsing \cite{Vinyals:2015:Grammar}, and semantic role
labelling \cite{zhou:2015}.
A key requirement for effectively training such models is an abundance of
supervised data.

In this paper we focus on learning mappings from input sequences $x$ to output
sequences $y$ in domains where the latter are easily obtained, but annotation in
the form of $(x,y)$ pairs is sparse or expensive to produce, and propose a novel
architecture that accommodates semi-supervised training on sequence transduction
tasks.
To this end, we augment the transduction objective ($x\mapsto y$) with an
autoencoding objective where the input sequence is treated as a latent
variable~($y\mapsto x\mapsto y$), enabling training from both labelled pairs and
unpaired output sequences. This is common in situations where we encode natural
language into a logical form governed by some grammar or database.

While such an autoencoder could in principle be constructed by stacking two
sequence transducers, modelling the latent variable as a series of discrete
symbols drawn from multinomial distributions creates serious computational
challenges, as it requires marginalising over the space of latent sequences
$\Sigma_x^*$. To avoid this intractable marginalisation, we introduce a novel
differentiable alternative for draws from a softmax which can be used with the
reparametrisation trick of \newcite{journals/corr/KingmaW13}. Rather than
drawing a discrete symbol in $\Sigma_x$ from a softmax, we draw a distribution
over symbols from a logistic-normal distribution at each time step. These serve
as continuous relaxations of discrete samples, providing a differentiable
estimator of the expected reconstruction log likelihood.

\begin{table*}[t]
  \centering
  \begin{tabular}{@{}l@{}l@{}}
    \toprule
    \textbf{Dataset} & \textbf{Example} \\
    \midrule
    \multirow{2}{*}{{\sc Geo}}
    & {\small what are the high points of states surrounding mississippi} \\
    & {\small answer(high\_point\_1(state(next\_to\_2(stateid('mississippi')))))} \\
    \midrule
    \multirow{2}{*}{\nlmaps\phantom{x}}
    & {\small Where are kindergartens in Hamburg?} \\
    & {\small
  query(area(keyval(`name',`Hamburg')),nwr(keyval(`amenity',`kindergarten')),qtype(latlong))} \\
    \midrule
    \multirow{2}{*}{\maze}
    & {\small turn right at the bench into the yellow tiled hall} \\
    & {\small $(1,6,90)$ FORWARD - FORWARD - RIGHT - STOP $(3,6,180)$} \\
    \bottomrule
  \end{tabular}
  \caption{Examples of natural language $x$ and logical form $y$
  from the three corpora and tasks used in this paper. Note that the \maze
  corpus requires additional information in order to map from the instruction to
  the action sequence.}
  \label{tab:tasks}
\end{table*}

\begin{figure*}[t!]\centering
  \includegraphics[width=\textwidth]{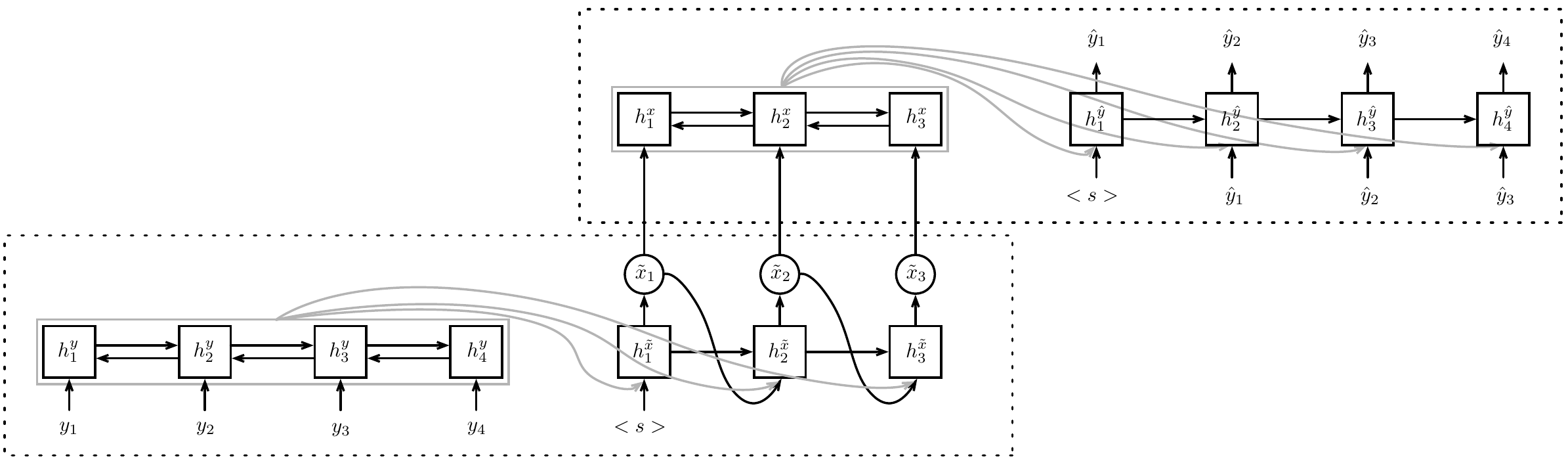}
  \caption{\xzyx model with attention-sequence-to-sequence encoder and
  decoder.  Circle nodes represent random variables.}
  \label{fig:model_overview}
\end{figure*}

We demonstrate the effectiveness of our proposed model on three semantic parsing
tasks: the \geoquery benchmark \cite{zelle:aaai96,Wong:2006:Geoquery}, the SAIL
maze navigation task \cite{MacMahon:2006:WTC:1597348.1597423} and the Natural
Language Querying corpus \cite{HaasRiezler:16} on OpenStreetMap. As part of our
evaluation, we introduce simple mechanisms for generating large amounts of
unsupervised training data for two of these tasks.

In most settings, the semi-supervised model outperforms the supervised model,
both when trained on additional generated data as well as on subsets of the
existing data.

\section{Model}
\label{sec:model}

Our sequential autoencoder is shown in Figure~\ref{fig:model_overview}.
At a high level, it can be seen as two sequence-to-sequence models with
attention \cite{bahdanau:2014:nmt} chained together. More precisely, the model
consists of four LSTMs~\cite{hochreiter:1997:lstm}, hence the name
\xzyx. The first, a bidirectional LSTM, encodes the sequence $y$; next, an LSTM
with stochastic output, described below, draws a sequence of distributions
$\tilde{x}$ over words in vocabulary $\Sigma_x$. The third LSTM encodes these
distributions for the last one to attend over and reconstruct $y$ as $\hat{y}$.
We now give the details of these parts.

\subsection{Encoding $y$}
The first LSTM of the encoder half of the model reads the sequence~$y$,
represented as a sequence of one-hot vectors over the vocabulary $\Sigma_y$,
using a bidirectional RNN into a sequence of vectors $h^{y}_{1:L_y}$ where
$L_y$ is the sequence length of $y$,
\begin{align}
  h^{y}_t &= \left(f^{\rightarrow}_y(y_t, h^{y,\rightarrow}_{t-1}) ;
                  f^{\leftarrow}_y(y_t, h^{y,\leftarrow}_{t+1})\right) ,
  \label{eq:enc.enc.ht}
\end{align}
where $f^{\rightarrow}_y,f^{\leftarrow}_y$ are non-linear functions applied at
each time step to the current token $y_t$ and their recurrent states
$h^{y,\rightarrow}_{t-1}$, $h^{y,\leftarrow}_{t+1}$, respectively.

Both the forward and backward functions project the one-hot vector into a dense
vector via an embedding matrix, which serves as input to an LSTM.

\subsection{Predicting a Latent Sequence $\tilde{x}$}

\begin{figure}[t]\centering
  \includegraphics[width=0.38\textwidth,trim={7.9cm 0 7.9cm 0},clip]{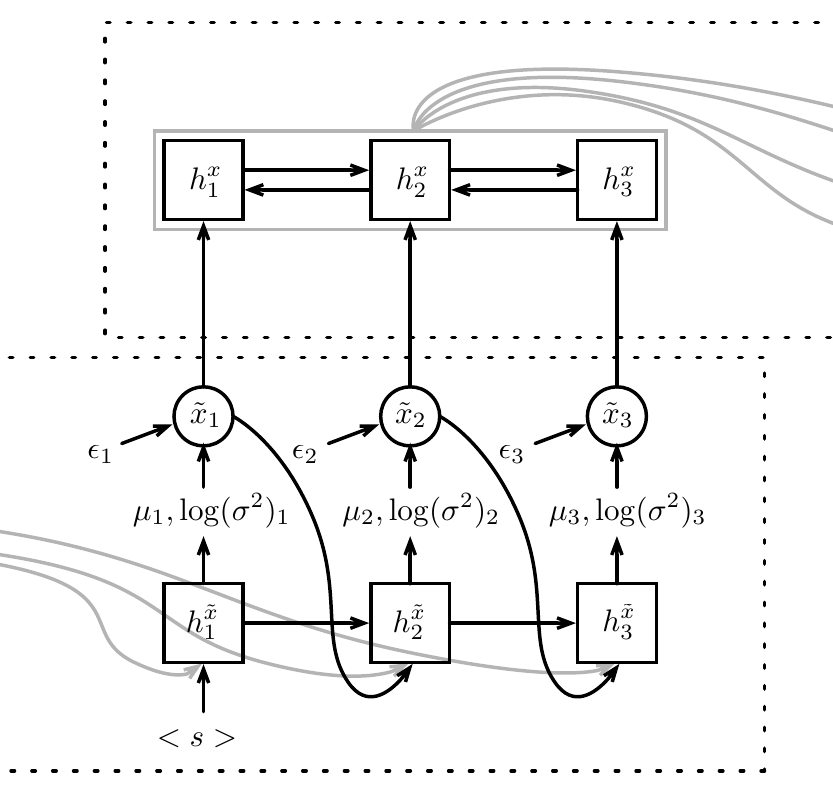}
  \caption{Unsupervised case of the \xzyx model.}
  \label{fig:model_unsup}
\end{figure}

Subsequently, we wish to predict $x$. Predicting a discrete sequence of
symbols through draws from multinomial distributions over a vocabulary
is not an option, as we would not be able to backpropagate through
this discrete choice. Marginalising over the possible latent strings or
estimating the gradient through na\"ive Monte Carlo methods would be a
prohibitively high variance process because the number of strings is exponential
in the maximum length (which we would have to manually specify) with the
vocabulary size as base. To allow backpropagation, we instead predict a sequence
of distributions $\tilde{x}$ over the symbols of $\Sigma_x$ with an RNN
attending over $h^{y} = h^{y}_{1:L_y}$, which will later serve to reconstruct
$y$:
\begin{align}
  \tilde{x} = q(x|y) = \prod^{L_x}_{t=1} q(\tilde{x}_t| \{\tilde{x}_1, \cdots, \tilde{x}_{t-1}\}, h^{y})
  \label{eq:enc.dec.prob}
\end{align}
where $q(x|y)$ models the mapping $y\mapsto x$. We define $q(\tilde{x}_t|
\{\tilde{x}_1, \cdots, \tilde{x}_{t-1}\}, h^{y})$ in the following way:

Let the vector $\tilde{x}_t$ be a distribution over the vocabulary $\Sigma_x$
drawn from a logistic-normal distribution\footnote{The logistic-normal
  distribution is the exponentiated and normalised (i.e.\ taking softmax) normal
  distribution.}, the parameters of which,
$\mu_t,\log(\sigma^2)_t\in\mathbb{R}^{|\Sigma_x|}$, are predicted by attending
by an LSTM attending over the outputs of the encoder
(Equation~\ref{eq:enc.dec.prob}), where $|\Sigma_x|$ is the size of the
vocabulary~$\Sigma_x$. The use of a logistic normal distribution serves to
regularise the model in the semi-supervised learning regime, which is described
at the end of this section.
Formally, this process, depicted in Figure~\ref{fig:model_unsup}, is as follows:
\begin{align}
  h^{\tilde{x}}_t &= f_{\tilde{x}}(\tilde{x}_{t-1}, h^{\tilde{x}}_{t-1}, h^{y}) \label{eq:enc.dec.ht.sup} \\
  \mu_t, \log(\sigma_t^2) &= l(h^{\tilde{x}}_t) \\
  \epsilon &\sim \mathcal{N}(0,I) \\
  \gamma_t &= \mu_t + \sigma_t \epsilon \\
  \tilde{x}_t &= \mathrm{softmax}(\gamma_t)
\end{align}
where the $f_{\tilde{x}}$ function is an LSTM and $l$ a linear transformation to
$\mathbb{R}^{2|\Sigma_x|}$. We use the reparametrisation trick
from~\newcite{journals/corr/KingmaW13} to draw from the logistic normal,
allowing us to backpropagate through the sampling process.

\subsection{Encoding $x$}

Moving on to the decoder part of our model, in the third LSTM, we
embed\footnote{Multiplying the distribution over words and an embedding matrix
  averages the word embedding of the entire vocabulary weighted by their
  probabilities.} and encode $\tilde{x}$:
\begin{align}
  h^{x}_t &= \left(f^{\rightarrow}_{x}(\tilde{x}_t, h^{x,\rightarrow}_{t-1}) ;
   f^{\leftarrow}_{x}(\tilde{x}_t, h^{x,\leftarrow}_{t+1})\right)
  \label{eq:dec.enc.ht}
\end{align}
When $x$ is observed, during supervised training and also when making
predictions, instead of the distribution $\tilde{x}$ we feed the one-hot encoded
$x$ to this part of the model.

\subsection{Reconstructing $y$}
In the final LSTM, we decode into $y$:
\begin{align}
p(\hat{y}|\tilde{x}) &= \prod^{L_y}_{t=1} p(\hat{y}_t| \{\hat{y}_1, \cdots,
\hat{y}_{t-1}\}, h^{\tilde{x}}) \label{eq:6}
\end{align}
Equation~\ref{eq:6} is implemented as an LSTM attending over $h^{\tilde{x}}$
producing a sequence of symbols $\hat{y}$ based on recurrent states
$h^{\hat{y}}$, aiming to reproduce input $y$:
\begin{align}
  h^{\hat{y}}_t &= f_{\hat{y}}(\hat{y}_{t-1}, h^{\hat{y}}_{t-1}, h^{\tilde{x}})
  \label{eq:dec.dec.ht} \\
  \hat{y}_t &\sim \mathrm{softmax}(l'(h^{\hat{y}}_t))
\end{align}
where $f_{\hat{y}}$ is the non-linear function, and the actual probabilities are
given by a softmax function after a linear transformation $l'$ of~$h^{\hat{y}}$.
At training time, rather than $\hat{y}_{t-1}$ we feed the ground truth
$y_{t-1}$.

\subsection{Loss function}

The complete model described in this section gives a reconstruction
function $y \mapsto \hat{y}$. We define a loss on this reconstruction
which accommodates the unsupervised case, where $x$ is not observed in
the training data, and the supervised case, where $(x,y)$ pairs are
available. Together, these allow us to train the \xzyx model in a
semi-supervised setting, which experiments will show provides some
benefits over a purely supervised training regime.

\paragraph{Unsupervised case}

When $x$ isn't observed, the loss we minimise during training is the
reconstruction loss on $y$, expressed as the negative log-likelihood
$NLL(\hat{y}, y)$ of the true labels $y$ relative to the predictions
$\hat{y}$. To this, we add as a regularising term the KL divergence
$\KL{q(\gamma|y)}{p(\gamma)}$ which effectively penalises the mean and
variance of $q(\gamma|y)$ from diverging from those of a prior
$p(\gamma)$, which we model as a diagonal Gaussian $\mathcal{N}(0,
I)$. This has the effect of smoothing the logistic normal distribution
from which we draw the distributions over symbols of $x$, guarding
against overfitting of the latent distributions over $x$ to symbols
seen in the supervised case discussed below.
The unsupervised loss is therefore formalised as
\begin{align}
  \mathcal{L}_{unsup} = NLL(\hat{y}, y) + \alpha \KL{q(\gamma|y)}{p(\gamma)}
\end{align}
with regularising factor $\alpha$ is tuned on validation, and
\begin{align}
  \KL{q(\gamma|y)}{p(\gamma)} = \sum_{i=1}^{L_x}{\KL{q(\gamma_i|y)}{p(\gamma)}}
\end{align}
We use a closed form of these individual KL divergences, described by \newcite{journals/corr/KingmaW13}.

\paragraph{Supervised case}

When $x$ is observed, we additionally minimise the prediction loss on
$x$, expressed as the negative log-likelihood $NLL(\tilde{x}, x)$ of the true
labels $x$ relative to the predictions $\tilde{x}$, and do not impose the KL
loss. The supervised loss is thus
\begin{align}
  \mathcal{L}_{sup} = NLL(\tilde{x}, x) + NLL(\hat{y}, y)
\end{align}

In both the supervised and unsupervised case, because of the continuous
relaxation on generating $\tilde{x}$ and the reparameterisation trick, the
gradient of the losses with regard to the model parameters is well defined
throughout \xzyx.

\paragraph{Semi-supervised training and inference} We train with a weighted
combination of the supervised and unsupervised losses described above. Once
trained, we simply use the $x \mapsto y$ decoder segment of the model to predict
$y$ from sequences of symbols $x$ represented as one-hot vectors. When the
decoder is trained without the encoder in a fully supervised manner, it serves
as our supervised sequence-to-sequence baseline model under the name \stos.

\section{Tasks and Data Generation}
\label{sec:data-generation}

We apply our model to three tasks outlined in this section. Moreover, we explain
how we generated additional unsupervised training data for two of these
tasks.
Examples from all datasets are in Table~\ref{tab:tasks}.

\subsection{GeoQuery}

The first task we consider is the prediction of a query on the \geocorpus which
is a frequently used benchmark for semantic parsing. The corpus contains 880
questions about US geography together with executable queries representing
those questions. We follow the approach established by
\newcite{zettlemoyer:2005} and split the corpus into 600 training and 280 test
cases. Following common practice, we augment the dataset by referring to the
database during training and test time. In particular, we use the database to
identify and anonymise variables (cities, states, countries and rivers)
following the method described in~\newcite{dong:2016}.

Most prior work on the \geocorpus relies on standard semantic parsing methods
together with custom heuristics or pipelines for this corpus. The recent paper
by \newcite{dong:2016} is of note, as it uses a sequence-to-sequence model for
training which is the unidirectional equivalent to \stos, and also to the decoder
part of our \xzyx network.

\subsection{Open Street Maps}

The second task we tackle with our model is the \nlmaps dataset by
\newcite{HaasRiezler:16}. The dataset contains 1,500 training and 880 testing
instances of
natural language questions with corresponding machine readable queries over the
geographical OpenStreetMap database. The dataset contains natural language
question in both English and German but we focus only on single language
semantic parsing, similar to the first task in \newcite{HaasRiezler:16}.
We use the data as it is, with the only pre-processing step being the
tokenization of both natural language and query form\footnote{We removed
quotes, added spaces around \texttt{(),} and separated the question mark from the
last word in each question.}.

\subsection{Navigational Instructions to Actions}

The \maze corpus and task were developed to train agents to follow free-form
navigational route instructions in a maze environment
\cite{MacMahon:2006:WTC:1597348.1597423,chen:aaai11}. It consists of a small
number of mazes containing features such as objects, wall and floor types.
These mazes come together with a large number of human instructions paired with
the required actions\footnote{There
are four actions: \texttt{LEFT}, \texttt{RIGHT}, \texttt{GO}, \texttt{STOP}.} to
reach the goal state described in those instructions.

We use the sentence-aligned version of the \maze route instruction dataset
containing 3,236 sentences \cite{chen:aaai11}.
Following previous work, we
accept an action sequence as correct if and only if the final position and
orientation exactly match those of the gold data.
We do not perform any pre-processing on this dataset.

\subsection{Data Generation}

As argued earlier, we are focusing on tasks where aligned data is sparse and
expensive to obtain, while it should be cheap to get unsupervised, monomodal
data. Albeit that is a reasonable assumption for real world data, the datasets
considered have no such component, thus the approach taken here is to generate
random database queries or maze paths, i.e. the machine readable side of the
data, and train a semi-supervised model. The alternative not explored here would
be to generate natural language questions or instructions instead, but that is
more difficult to achieve without human intervention.  For this reason, we
generate the machine readable side of the data for \geoquery and \maze
tasks\footnote{Our randomly generated unsupervised datasets can be downloaded
from \url{http://deepmind.com/publications}}.

For \geoquery, we fit a 3-gram Kneser-Ney \cite{chen1999empirical} model to the
queries in the training set and sample about 7 million queries from it. We
ensure that the sampled queries are different from the training queries, but do
not enforce validity. This intentionally simplistic approach is to demonstrate
the applicability of our model.

The \maze dataset has only three mazes. We added a fourth one and over 150k
random paths, including duplicates. The new maze is larger ($21 \times 21$
grid) than the existing ones, and seeks to approximately replicate the key
statistics of the other three mazes (maximum corridor length, distribution of
objects, etc). Paths within that maze are created by randomly sampling start and
end positions.

\section{Experiments}
\label{sec:experiments}

We evaluate our model on the three tasks in multiple settings. First, we
establish a supervised baseline to compare the \stos model with prior work.
Next, we train our \xzyx model in a semi-supervised setting on the entire
dataset with the additional monomodal training data described in the previous
section.

Finally, we perform an ``ablation'' study where we discard some of the training
data and compare \stos to \xzyx. \stos is trained solely on the reduced data in
a supervised manner, while \xzyx is once again trained semi-supervised on the
same reduced data plus the machine readable part of the discarded data
(\xzyx{}-) or on the extra generated data (\xzyx{}+).

\paragraph{Training}

We train the model using standard gradient descent methods. As none of the
datasets used here contain development sets, we tune hyperparameters by
cross-validating on the training data. In the case of the \maze corpus we train
on three folds (two mazes for training and validation, one for test each) and
report weighted results across the folds following prior work
\cite{mei2016navigational}.

\subsection{GeoQuery}

\begin{table}[t]
  \centering
  \begin{tabular}{@{}lr@{}}
    \toprule
    Model & Accuracy \\
    \midrule
    \newcite{zettlemoyer:2005} & 79.3 \\
    \newcite{zettlemoyer:2007:ccg} & 86.1 \\
    \newcite{liang2013learning} & 87.9 \\
    \newcite{Kwiatkowski:2011:LGCCG} & 88.6 \\
    \newcite{zhao2014type} & 88.9 \\
    \newcite{kwiatkowski2013scaling} & 89.0 \\
    \midrule
    \newcite{dong:2016} & 84.6 \\
    \newcite{jia2016recombination}\protect\tablefootnote{%
      \newcite{jia2016recombination}
      used hand crafted grammars to generate additional supervised training data.
      } & 89.3 \\
    \midrule
    \stos & 86.5 \\
    \xzyxqlq & 87.3 \\
    \bottomrule
  \end{tabular}
  \caption{Non-neural and neural model results on \geoquery using the
  train/test split from \protect\cite{zettlemoyer:2005}.
  }
  \label{tab:geoquery}
\end{table}

The evaluation metric for \geoquery is the accuracy of exactly predicting the
machine readable query. As results in Table~\ref{tab:geoquery} show, our
supervised \stos baseline model performs slightly better than the comparable
model by \newcite{dong:2016}. The semi-supervised \xzyxqlq model with the
additional generated queries improves on it further.

The ablation study in Table~\ref{tab:geoquery-ablation} demonstrates a widening
gap between supervised and semi-supervised as the amount of labelled training
data gets smaller. This suggests that our model can leverage unlabelled data
even when only small amount of labelled data is available.

\begin{table}[t]
  \centering
  \begin{tabular}{@{}lrrr@{}}
    \toprule
    Sup. data & \stos & \xzyxqlq{}- & \xzyxqlq{}+ \\
    \midrule
    5\%       &  21.9 & 30.1 & 26.2 \\
    10\%      &  39.7 & 42.1 & 42.1 \\
    25\%      &  62.4 & 70.4 & 67.1 \\
    50\%      &  80.3 & 81.2 & 80.4 \\
    75\%      &  85.3 & 84.1 & 85.1 \\
    100\%     &  86.5 & 86.5 & 87.3 \\
    \bottomrule
  \end{tabular}
  \caption{Results of the \geoquery ablation study.}
  \label{tab:geoquery-ablation}
\end{table}

\subsection{Open Street Maps}

We report results for the \nlmaps corpus in Table~\ref{tab:nlmaps}, comparing
the supervised \stos model to the results posted by \newcite{HaasRiezler:16}.
While their model used a semantic parsing pipeline including alignment,
stemming, language modelling and CFG inference, the strong performance of the
\stos model demonstrates the strength of fairly vanilla attention-based
sequence-to-sequence models.
It should be pointed out that the previous work reports the number of correct
answers when queries were executed against the dataset, while we evaluate on the
strict accuracy of the generated queries. While we expect these numbers to
be nearly equivalent, our evaluation is strictly harder as it does not allow for
reordering of query arguments and similar relaxations.

We investigate the \xzyxqlq model only via the ablation study in
Table~\ref{tab:nlmapsabl} and find little gain through the semi-supervised
objective. Our attempt at cheaply generating unsupervised data for this task was
not successful, likely due to the complexity of the underlying database.

\begin{table}[t]
  \centering
  \begin{tabular}{@{}lr@{}}
    \toprule
    Model & Accuracy \\
    \midrule
    \newcite{HaasRiezler:16} & 68.30 \\
    \stos & 78.03 \\
    \bottomrule
  \end{tabular}
  \caption{Results on the \nlmaps corpus.}\label{tab:nlmaps}
\end{table}

\begin{table}[t]
  \centering
  \begin{tabular}{@{}lrr@{}}
    \toprule
    Sup. data & \stos & \xzyxqlq{}- \\
    \midrule
     5\% &   3.22 & 3.74  \\
    10\% &  17.61 & 17.12 \\
    25\% &  33.74 & 33.50 \\
    50\% &  49.52 & 53.72 \\
    75\% &  66.93 & 66.45 \\
    100\% & 78.03 & 78.03 \\
    \bottomrule
  \end{tabular}
  \caption{Results of the \nlmaps ablation study.}
  \label{tab:nlmapsabl}
\end{table}

\subsection{Navigational Instructions to Actions}

\paragraph{Model extension}
The experiments for the \maze task differ slightly from the other two tasks in
that the language input does not suffice for choosing an action.
While a simple instruction such as \textit{`turn left'} can easily be
translated into the action sequence \texttt{LEFT-STOP}, more complex
instructions such as \textit{`Walk forward until you see a lamp'} require
knowledge of the agent's position in the maze.

To accomplish this we modify the model as follows. First, when encoding action
sequences, we concatenate each action with a representation of the maze
at the given position, representing the maze-state akin to
\newcite{mei2016navigational} with a bag-of-features vector. Second, when
decoding action sequences, the RNN outputs an action which is used to update the
agent's position and the representation of that new position is fed into the RNN
as its next input.

\paragraph{Training regime}
We cross-validate over the three mazes in the dataset and report
overall results weighted by test size (cf.
\newcite{mei2016navigational}).
Both our supervised and semi-supervised model perform worse than the
state-of-the-art (see Table \ref{tab:maze}), but the latter enjoys a comfortable
margin
over the former. As the \stos model broadly reimplements the work of
\newcite{mei2016navigational}, we put the discrepancy in performance down to the
particular design choices that we did not follow in order to keep the model here
as general as possible and comparable across tasks.

The ablation studies (Table \ref{tab:mazeabl}) show little gain for the
semi-supervised approach when only using data from the original
training set, but substantial improvement with the additional unsupervised data.

\begin{table}[t]
  \centering
  \begin{tabular}{@{}lr@{}}
    \toprule
    Model & Accuracy \\
    \midrule
    \newcite{chen:aaai11} & 54.40 \\
    \newcite{kim.emnlp12} & 57.22 \\
    \newcite{Andreas-Klein:2015:AlignmentSemantics} & 59.60 \\
    \newcite{kim-mooney:2013:ACL2013} & 62.81 \\
    \newcite{artzi-das-petrov:2014:EMNLP2014} & 64.36 \\
    \newcite{artzi-zettlemoyer:2013:TACL} & 65.28 \\
    \midrule
    \newcite{mei2016navigational} & 69.98 \\
    \midrule
    \stos & 58.60 \\
    \xzyxqlq & 63.25 \\
    \bottomrule
  \end{tabular}
  \caption{Results on the \maze corpus.}\label{tab:maze}
\end{table}

\begin{table}[t]
  \centering
  \begin{tabular}{@{}lrrr@{}}
    \toprule
    Sup. data & \stos & \xzyxqlq{}- & \xzyxqlq{}+\\
    \midrule
    5\% &  37.79 & 41.48 & 43.44 \\
    10\% &  40.77 & 41.26 & 48.67 \\
    25\% &  43.76 & 43.95 & 51.19 \\
    50\% &  48.01 & 49.42 & 55.97 \\
    75\% &  48.99 & 49.20 & 57.40 \\
    100\% & 49.49 & 49.49 & 58.28 \\
    \bottomrule
  \end{tabular}
  \caption{Results of the \maze ablation study. Results are from models trained
    on \textit{L} and \textit{Jelly} maps, tested on \textit{Grid} only, hence
    the discrepancy between the 100\% result and \stos in Table
    \protect\ref{tab:maze}.}\label{tab:mazeabl}
\end{table}

\section{Discussion}

\begin{table*}[t]
  \centering
  \begin{tabular}{@{}ll@{}}
    \toprule
     \textbf{Input from unsupervised data ($y$)}
     & \textbf{Generated latent representation ($x$)} \\
    \midrule
     {\small answer smallest city loc\_2 state stateid \_STATE\_}
     & {\small what is the smallest city in the state of \_STATE\_ $<$/S$>$} \\
     {\small answer city loc\_2 state next\_to\_2 stateid \_STATE\_}
     & {\small what are the cities in states which border \_STATE\_ $<$/S$>$} \\
    \midrule
     {\small answer mountain loc\_2 countryid \_COUNTRY\_}
     & {\small what is the lakes in \_COUNTRY\_ $<$/S$>$} \\
     {\small answer state next\_to\_2 state all }
     & {\small which states longer states show peak states to $<$/S$>$} \\
    \bottomrule
  \end{tabular}
  \caption{Positive and negative examples of latent language together with the randomly generated
  logical form from the unsupervised part of the \geoquery training. Note that
the natural language ($x$) does not occur anywhere in the training data in this
form.}
\label{tab:qualitative}\vspace{-5pt}
\end{table*}

\paragraph{Supervised training}

The prediction accuracies of our supervised baseline \stos model are mixed with
respect to prior results on their respective tasks.
For \geoquery, \stos performs significantly better than the most similar model
from the literature \cite{dong:2016}, mostly due to the fact that $y$ and $x$
are encoded with bidirectional LSTMs. With a unidirectional LSTM we get similar
results to theirs.

On the \maze corpus, \stos performs worse than the state of the art. As the
models are broadly equivalent we attribute this difference to a number of
task-specific choices and optimisations\footnote{In particular we don't use beam
search and ensembling.}
made in \newcite{mei2016navigational} which we did not
reimplement for the sake of using a common model across all three tasks.

For \nlmaps, \stos performs much better than the state-of-the-art, exceeding the
previous best result by 11\% despite a very simple tokenization method and a
lack of any form of entity anonymisation.

\paragraph{Semi-supervised training}

In both the case of \geoquery and the \maze task we found the
semi-supervised model to convincingly outperform the fully supervised model.
The effect was particularly notable in the case of the \maze corpus, where performance
increased from $58.60\%$ accuracy to $63.25\%$ (see Table \ref{tab:maze}).
It is worth remembering that the supervised training regime consists of three
folds of tuning on two maps with subsequent testing on the third map, which
carries a risk of overfitting to the training maps. The introduction of the
fourth unsupervised map clearly mitigates this effect.
Table \ref{tab:qualitative} shows some examples of unsupervised logical forms
being transformed into natural language, which demonstrate how the model can
learn to sensibly ground unsupervised data.

\paragraph{Ablation performance}

The experiments with additional unsupervised data prove the feasibility of
our approach and clearly demonstrate the usefulness of the \xzyxqlq model for
the general class of sequence-to-sequence tasks where supervised data is hard to
come by. To analyse the model further, we also look at the performance of both
\stos and \xzyxqlq when reducing the amount of supervised training data
available to the model. We compare three settings: the supervised \stos model
with reduced training data, \xzyxqlq{}- which uses the removed training data in
an unsupervised fashion (throwing away the natural language) and
\xzyxqlq{}+ which uses the randomly generated unsupervised data described in
Section \ref{sec:data-generation}.
The \stos model behaves as expected on all three tasks, its performance
dropping with the size of the training data.
The performance of \xzyxqlq{}- and \xzyxqlq{}+ requires more
analysis.

In the case of \geoquery, having unlabelled data from the true distribution
(\xzyxqlq{}-) is a good thing when there is enough of it, as clearly seen when
only 5\% of the original dataset is used for supervised training and the
remaining 95\% is used for unsupervised training. The gap shrinks as the amount
of supervised data is increased, which is as expected.
On the other hand, using a large amount of extra, generated data from an
approximating distribution (\xzyxqlq{}+) does not help as much initially when
compared with the unsupervised data from the true distribution. However, as the
size of the unsupervised dataset in \xzyxqlq{}- becomes the bottleneck this gap
closes and eventually the model trained on the extra data achieves higher
accuracy.

For the \maze task the semi-supervised models do better than the
supervised results throughout, with the model trained on randomly generated
additional data consistently outperforming the model trained only on the
original data. This gives further credence to the risk of overfitting to the
training mazes already mentioned above.

Finally, in the case of the \nlmaps corpus, the semi-supervised approach does not
appear to help much at any point during the ablation. These indistinguishable
results are likely due to the task's complexity, causing the ablation
experiments to either have to little supervised data to sufficiently ground the
latent space to make use of the unsupervised data, or in the higher percentages
then too little unsupervised data to meaningfully improve the model.

\section{Related Work}

\paragraph{Semantic parsing}

The tasks in this paper all broadly belong to the domain of semantic
parsing, which describes the process of mapping natural language to a formal
representation of its meaning. This is extended in the \maze
navigation task, where the formal representation is a function of both the
language instruction and a given environment.

Semantic parsing is a well-studied problem with numerous approaches including
inductive logic programming \cite{zelle:aaai96}, string-to-tree
\cite{galley-EtAl:2004:HLTNAACL} and string-to-graph \cite{Jones:2012:Hyperedge}
transducers, grammar induction
\cite{Kwiatkowski:2011:LGCCG,artzi-zettlemoyer:2013:TACL,Reddy:2014:SMwithout}
or machine translation \cite{Wong:2006:Geoquery,Andreas:2013:SPasMT}.

While a large number of relevant literature focuses on defining the grammar of
the logical forms \cite{zettlemoyer:2005}, other models learn purely from
aligned pairs of text and logical form \cite{Berant:2014:SPvPP}, or from more
weakly supervised signals such as question-answer pairs together with a database
\cite{Liang:2011:Depbased}.
Recent work of \newcite{jia2016recombination} induces a synchronous
context-free grammar and generates additional training examples~$(x,y)$, which
is one way to address data scarcity issues.
The semi-supervised setup proposed here offers an alternative solution to this
issue.

\paragraph{Discrete autoencoders}

Very recently there has been some related work on discrete autoencoders for
natural language processing \cite[\textit{i.a.}]{Suster:2016,Marcheggiani:2016:DiscVarAuto}
This work presents a first approach to using effectively discretised sequential
information as the latent representation without resorting to draconian
assumptions \cite{wammar:2014} to make marginalisation tractable. While our
model is not exactly marginalisable either, the continuous relaxation makes
training far more tractable. A related idea was recently presented in
\newcite{deep-fusion-2015}, who use monolingual data to improve machine
translation by fusing a sequence-to-sequence model and a language model.

\section{Conclusion}

We described a method for augmenting a supervised sequence transduction
objective with an autoencoding objective, thereby enabling semi-supervised
training where previously a scarcity of aligned data might have held back model
performance.  Across multiple semantic parsing tasks we demonstrated the
effectiveness of this approach, improving model performance by training on
randomly generated unsupervised data in addition to the original data.

Going forward it would be interesting to further analyse the effects of sampling
from a logistic-normal distribution as opposed to a softmax in order to better
understand how this impacts the distribution in the latent space.  While we
focused on tasks with little supervised data and additional unsupervised data in
$y$, it would be straightforward to reverse the model to train it with
additional labelled data in $x$, i.e. on the natural language side. A natural
extension would also be a formulation where semi-supervised training was
performed in both $x$ and $y$. For instance, machine translation lends itself to
such a formulation where for many language pairs parallel data may be scarce
while there is an abundance of monolingual data.

\bibliographystyle{emnlp2016}
\bibliography{paper}

\begin{thebibliography}{}

\bibitem[\protect\citename{Ammar \bgroup et al.\egroup }2014]{wammar:2014}
Waleed Ammar, Chris Dyer, and Noah~A. Smith.
\newblock 2014.
\newblock {Conditional Random Field Autoencoders for Unsupervised Structured
  Prediction}.
\newblock In {\em Proceedings of NIPS}.

\bibitem[\protect\citename{Andreas and
  Klein}2015]{Andreas-Klein:2015:AlignmentSemantics}
Jacob Andreas and Dan Klein.
\newblock 2015.
\newblock {Alignment-based Compositional Semantics for Instruction Following}.
\newblock In {\em Proceedings of EMNLP}, September.

\bibitem[\protect\citename{Andreas \bgroup et al.\egroup
  }2013]{Andreas:2013:SPasMT}
Jacob Andreas, Andreas Vlachos, and Stephen Clark.
\newblock 2013.
\newblock {Semantic Parsing as Machine Translation}.
\newblock In {\em Proceedings of ACL}, August.

\bibitem[\protect\citename{Artzi and
  Zettlemoyer}2013]{artzi-zettlemoyer:2013:TACL}
Yoav Artzi and Luke Zettlemoyer.
\newblock 2013.
\newblock {Weakly Supervised Learning of Semantic Parsers for Mapping
  Instructions to Actions}.
\newblock {\em Transactions of the Association for Computational Linguistics},
  1(1):49--62.

\bibitem[\protect\citename{Artzi \bgroup et al.\egroup
  }2014]{artzi-das-petrov:2014:EMNLP2014}
Yoav Artzi, Dipanjan Das, and Slav Petrov.
\newblock 2014.
\newblock {Learning Compact Lexicons for CCG Semantic Parsing}.
\newblock In {\em Proceedings of EMNLP}, October.

\bibitem[\protect\citename{Bahdanau \bgroup et al.\egroup
  }2015]{bahdanau:2014:nmt}
Dzmitry Bahdanau, Kyunghyun Cho, and Yoshua Bengio.
\newblock 2015.
\newblock {Neural Machine Translation by Jointly Learning to Align and
  Translate}.
\newblock In {\em Proceedings of ICLR}.

\bibitem[\protect\citename{Berant and Liang}2014]{Berant:2014:SPvPP}
Jonathan Berant and Percy Liang.
\newblock 2014.
\newblock {Semantic Parsing via Paraphrasing}.
\newblock In {\em Proceedings of ACL}, June.

\bibitem[\protect\citename{Chen and Goodman}1999]{chen1999empirical}
Stanley~F Chen and Joshua Goodman.
\newblock 1999.
\newblock {An empirical study of smoothing techniques for language modeling}.
\newblock {\em Computer Speech \& Language}, 13(4):359--393.

\bibitem[\protect\citename{Chen and Mooney}2011]{chen:aaai11}
David~L. Chen and Raymond~J. Mooney.
\newblock 2011.
\newblock {Learning to Interpret Natural Language Navigation Instructions from
  Observations}.
\newblock In {\em Proceedings of AAAI}, August.

\bibitem[\protect\citename{Dong and Lapata}2016]{dong:2016}
Li~Dong and Mirella Lapata.
\newblock 2016.
\newblock {Language to Logical Form with Neural Attention}.
\newblock {\em arXiv preprint arXiv:1601.01280}.

\bibitem[\protect\citename{Galley \bgroup et al.\egroup
  }2004]{galley-EtAl:2004:HLTNAACL}
Michel Galley, Mark Hopkins, Kevin Knight, and Daniel Marcu.
\newblock 2004.
\newblock {What's in a translation rule?}
\newblock In {\em Proceedings of HLT-NAACL}, May.

\bibitem[\protect\citename{G{\"{u}}l{\c{c}}ehre \bgroup et al.\egroup
  }2015]{deep-fusion-2015}
{\c{C}}aglar G{\"{u}}l{\c{c}}ehre, Orhan Firat, Kelvin Xu, Kyunghyun Cho,
  Lo{\"{\i}}c Barrault, Huei{-}Chi Lin, Fethi Bougares, Holger Schwenk, and
  Yoshua Bengio.
\newblock 2015.
\newblock {On Using Monolingual Corpora in Neural Machine Translation}.
\newblock {\em arXiv preprint arXiv:1503.03535}.

\bibitem[\protect\citename{Haas and Riezler}2016]{HaasRiezler:16}
Carolin Haas and Stefan Riezler.
\newblock 2016.
\newblock A corpus and semantic parser for multilingual natural language
  querying of openstreetmap.
\newblock In {\em Proceedings of NAACL}, June.

\bibitem[\protect\citename{Hochreiter and
  Schmidhuber}1997]{hochreiter:1997:lstm}
Sepp Hochreiter and J\"{u}rgen Schmidhuber.
\newblock 1997.
\newblock {Long Short-Term Memory}.
\newblock {\em Neural Computation}, 9(8):1735--1780, November.

\bibitem[\protect\citename{Jia and Liang}2016]{jia2016recombination}
Robin Jia and Percy Liang.
\newblock 2016.
\newblock Data recombination for neural semantic parsing.
\newblock In {\em Association for Computational Linguistics (ACL)}.

\bibitem[\protect\citename{Jones \bgroup et al.\egroup
  }2012]{Jones:2012:Hyperedge}
Bevan Jones, Jacob Andreas, Daniel Bauer, Karl~Moritz Hermann, and Kevin
  Knight.
\newblock 2012.
\newblock {Semantics-Based Machine Translation with Hyperedge Replacement
  Grammars}.
\newblock In {\em Proceedings of COLING 2012}, December.

\bibitem[\protect\citename{Kim and Mooney}2012]{kim.emnlp12}
Joohyun Kim and Raymond~J. Mooney.
\newblock 2012.
\newblock {Unsupervised PCFG Induction for Grounded Language Learning with
  Highly Ambiguous Supervision}.
\newblock In {\em Proceedings of EMNLP-CoNLL}, July.

\bibitem[\protect\citename{Kim and Mooney}2013]{kim-mooney:2013:ACL2013}
Joohyun Kim and Raymond Mooney.
\newblock 2013.
\newblock {Adapting Discriminative Reranking to Grounded Language Learning}.
\newblock In {\em Proceedings of ACL}, August.

\bibitem[\protect\citename{Kingma and Welling}2014]{journals/corr/KingmaW13}
Diederik~P. Kingma and Max Welling.
\newblock 2014.
\newblock {Auto-Encoding Variational Bayes.}
\newblock In {\em Proceedings of ICLR}.

\bibitem[\protect\citename{Kwiatkowski \bgroup et al.\egroup
  }2011]{Kwiatkowski:2011:LGCCG}
Tom Kwiatkowski, Luke Zettlemoyer, Sharon Goldwater, and Mark Steedman.
\newblock 2011.
\newblock {Lexical Generalization in CCG Grammar Induction for Semantic
  Parsing}.
\newblock In {\em Proceedings of EMNLP}.

\bibitem[\protect\citename{Kwiatkowski \bgroup et al.\egroup
  }2013]{kwiatkowski2013scaling}
Tom Kwiatkowski, Eunsol Choi, Yoav Artzi, and Luke Zettlemoyer.
\newblock 2013.
\newblock {Scaling semantic parsers with on-the-fly ontology matching}.
\newblock In {\em In Proceedings of EMNLP}. Citeseer.

\bibitem[\protect\citename{Liang \bgroup et al.\egroup
  }2011]{Liang:2011:Depbased}
Percy Liang, Michael~I. Jordan, and Dan Klein.
\newblock 2011.
\newblock {Learning Dependency-based Compositional Semantics}.
\newblock In {\em Proceedings of the ACL-HLT}.

\bibitem[\protect\citename{Liang \bgroup et al.\egroup
  }2013]{liang2013learning}
Percy Liang, Michael~I Jordan, and Dan Klein.
\newblock 2013.
\newblock {Learning dependency-based compositional semantics}.
\newblock {\em Computational Linguistics}, 39(2):389--446.

\bibitem[\protect\citename{MacMahon \bgroup et al.\egroup
  }2006]{MacMahon:2006:WTC:1597348.1597423}
Matt MacMahon, Brian Stankiewicz, and Benjamin Kuipers.
\newblock 2006.
\newblock {Walk the Talk: Connecting Language, Knowledge, and Action in Route
  Instructions}.
\newblock In {\em Proceedings of AAAI}.

\bibitem[\protect\citename{Marcheggiani and
  Titov}2016]{Marcheggiani:2016:DiscVarAuto}
Diego Marcheggiani and Ivan Titov.
\newblock 2016.
\newblock Discrete-state variational autoencoders for joint discovery and
  factorization of relations.
\newblock {\em Transactions of ACL}.

\bibitem[\protect\citename{Mei \bgroup et al.\egroup
  }2016]{mei2016navigational}
Hongyuan Mei, Mohit Bansal, and Matthew~R. Walter.
\newblock 2016.
\newblock {Listen, Attend, and Walk: Neural Mapping of Navigational
  Instructions to Action Sequences}.
\newblock In {\em Proceedings of AAAI}.

\bibitem[\protect\citename{Reddy \bgroup et al.\egroup
  }2014]{Reddy:2014:SMwithout}
Siva Reddy, Mirella Lapata, and Mark Steedman.
\newblock 2014.
\newblock {Large-scale Semantic Parsing without Question-Answer Pairs}.
\newblock {\em Transactions of the Association for Computational Linguistics},
  2:377--392.

\bibitem[\protect\citename{Suster \bgroup et al.\egroup }2016]{Suster:2016}
Simon Suster, Ivan Titov, and Gertjan van Noord.
\newblock 2016.
\newblock {Bilingual Learning of Multi-sense Embeddings with Discrete
  Autoencoders}.
\newblock {\em CoRR}, abs/1603.09128.

\bibitem[\protect\citename{Vinyals \bgroup et al.\egroup
  }2015]{Vinyals:2015:Grammar}
Oriol Vinyals, \L{}ukasz Kaiser, Terry Koo, Slav Petrov, Ilya Sutskever, and
  Geoffrey Hinton.
\newblock 2015.
\newblock {Grammar as a Foreign Language}.
\newblock In {\em Proceedings of NIPS}.

\bibitem[\protect\citename{Wong and Mooney}2006]{Wong:2006:Geoquery}
Yuk~Wah Wong and Raymond~J. Mooney.
\newblock 2006.
\newblock {Learning for Semantic Parsing with Statistical Machine Translation}.
\newblock In {\em Proceedings of NAACL}.

\bibitem[\protect\citename{Zelle and Mooney}1996]{zelle:aaai96}
John~M. Zelle and Raymond~J. Mooney.
\newblock 1996.
\newblock {Learning to Parse Database Queries using Inductive Logic
  Programming}.
\newblock In {\em Proceedings of AAAI/IAAI}, pages 1050--1055, August.

\bibitem[\protect\citename{Zettlemoyer and Collins}2005]{zettlemoyer:2005}
Luke~S. Zettlemoyer and Michael Collins.
\newblock 2005.
\newblock {Learning to Map Sentences to Logical Form: Structured Classification
  with Probabilistic Categorial Grammars.}
\newblock In {\em UAI}, pages 658--666. AUAI Press.

\bibitem[\protect\citename{Zettlemoyer and Collins}2007]{zettlemoyer:2007:ccg}
Luke Zettlemoyer and Michael Collins.
\newblock 2007.
\newblock {Online Learning of Relaxed CCG Grammars for Parsing to Logical
  Form}.
\newblock In {\em Proceedings of EMNLP-CoNLL}, June.

\bibitem[\protect\citename{Zhao and Huang}2014]{zhao2014type}
Kai Zhao and Liang Huang.
\newblock 2014.
\newblock {Type-driven incremental semantic parsing with polymorphism}.
\newblock {\em arXiv preprint arXiv:1411.5379}.

\bibitem[\protect\citename{Zhou and Xu}2015]{zhou:2015}
Jie Zhou and Wei Xu.
\newblock 2015.
\newblock {End-to-end Learning of Semantic Role Labeling Using Recurrent Neural
  Networks}.
\newblock In {\em Proceedings of ACL}.

\end{thebibliography}

\end{document}